\pdfoutput=1

\documentclass[11pt]{article}

\usepackage{ACL2023}

\usepackage{times}
\usepackage{latexsym}

\usepackage[T1]{fontenc}

\usepackage[utf8]{inputenc}

\usepackage{microtype}

\usepackage{inconsolata}
\usepackage{subcaption}
\usepackage{tabularx}
\usepackage{multirow}
\usepackage{graphicx}
\usepackage{booktabs}
\usepackage[symbol]{footmisc}

\title{AppAgent: Multimodal Agents as Smartphone Users}

\author{Chi Zhang$^{\ast}$ 
\quad
Zhao Yang$^{\ast}$ 
\quad Jiaxuan Liu$^{\ast}$ \quad Yucheng Han \quad Xin Chen \\ 
\bf
 \quad Zebiao Huang \quad Bin Fu   \quad Gang Yu$^{\dagger}$ \vspace{0.3em} \\
{ Tencent}  \\
{\normalsize \{johnczhang,~jayzyang,~jiaxuanliu,~yuchenghan,~shingxchen,~zebiaohuang,~brianfu,~skicyyu\}@tencent.com} \\
\url{https://appagent-official.github.io/}
}

\begin{document}

\twocolumn[{
\renewcommand\twocolumn[1][]{#1}
\maketitle
\centering
\vspace{-0.5cm}
\includegraphics[width=.95\linewidth]{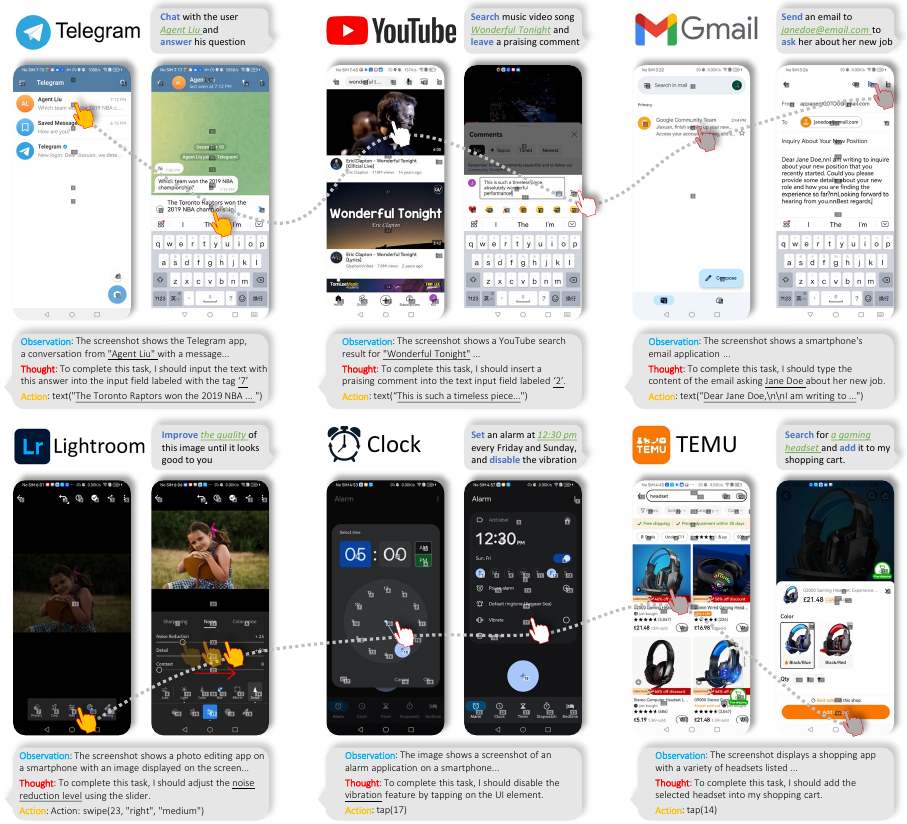}
\captionsetup{type=figure}
\caption{\textbf{Diverse applications of our multimodal agent framework for smartphone App operation.} We evaluate the effectiveness of our agent model on 50 tasks across 10 different Apps, highlighting its adaptability and effectiveness in a real-world context.}
\vspace{0.2cm}
\label{fig:teaser}
}]

\footnotetext[1]{Equal contributions.}
\footnotetext[2]{Corresponding Author.}

\begin{abstract}
Recent advancements in large language models (LLMs) have led to the creation of intelligent agents capable of performing complex tasks. This paper introduces a novel LLM-based multimodal agent framework designed to operate smartphone applications.
 Our framework enables the agent to operate smartphone applications through a simplified action space, mimicking human-like interactions such as tapping and swiping. This novel approach bypasses the need for system back-end access, thereby broadening its applicability across diverse apps.
Central to our agent's functionality is its innovative learning method. The agent learns to navigate and use new apps either through autonomous exploration or by observing human demonstrations. This process generates a knowledge base that the agent refers to for executing complex tasks across different applications.
To demonstrate the practicality of our agent, we conducted extensive testing over 50 tasks in 10 different applications, including social media, email, maps, shopping, and sophisticated image editing tools. The results affirm our agent's proficiency in handling a diverse array of high-level tasks.

\end{abstract}

\section{Introduction}
The emergence of large language models (LLMs), such as ChatGPT~\cite{chatgpt} and GPT-4~\cite{gpt4}, marks a significant milestone in the field of artificial intelligence and natural language processing. These advanced models represent a fundamental change in how machines understand and generate human language, exhibiting a level of sophistication and versatility previously unattainable.
One of the most exciting developments in this field is the capability of LLMs to function not just as language processors, but as agents capable of performing complex tasks. This evolution is evident in initiatives such as AutoGPT~\cite{yang2023autogpt} and MetaGPT~\cite{hong2023metagpt}, which showcase the practical applications of LLMs in tasks requiring advanced cognitive functions like reasoning, planning, and collaboration. The significance of these developments cannot be overstated, as they extend the utility of LLMs beyond simple language tasks, revolutionizing various aspects of technology and daily life.

However, a key limitation of these LLM-based agents has been their reliance solely on text-based information. This restriction has historically curtailed their perception and interaction with their environment.
 The introduction of models equipped with vision capabilities, such as the latest iteration of GPT-4, marks a pivotal breakthrough. By integrating the ability to process and interpret visual information, these models can now understand aspects of their surroundings that are difficult or impossible to convey through text alone. This extended capability enables LLMs to interpret context, recognize patterns, and respond to visual cues, thus providing a more holistic and interactive experience with the world.

In our work, we focus on building a multimodal agent leveraging the vision capabilities of multimodal large language models to undertake tasks previously unachievable by text-only agents. 
In particular, we explore an interesting but challenging application that builds an agent to operate any smartphone application (App) in the mobile operating system.
\emph{Our approach differs significantly from existing intelligent phone assistants like Siri, which operate through system back-end access and function calls.} Instead, our agent interacts with smartphone apps in a human-like manner, using low-level operations such as tapping and swiping on the graphical user interface (GUI). The proposed agent offers multiple advantages. Firstly, it eliminates the need for system back-end access, making our agent universally applicable across various applications. Additionally, this approach enhances security and privacy, as the agent does not require deep system integration. Furthermore, by operating on the GUI level, our agent can adapt to changes in app interfaces and updates, ensuring long-term applicability and flexibility.

However, creating a multimodal agent capable of operating diverse smartphone apps presents significant challenges. 
Existing research indicates that adapting current models for embodied tasks necessitates extensive training data, and collecting a large dataset of app demonstrations for training is a formidable task.
Moreover, \emph{different apps have unique GUIs with varying icon meanings and operational logic}, and it remains uncertain whether these adapted models can effectively generalize to unseen apps.

In this paper, we introduce a multimodal agent framework aimed at operating any smartphone app like human users. The learning of our framework involves an exploration phase where the agent interacts autonomously with apps through a set of pre-defined actions and learns from their outcomes. 
 These interactions are documented, which assists the agent in navigating and operating the apps. This learning process can be accelerated by observing a few human demonstrations. 
 Following this exploratory phase, the agent can operate the app by consulting the constructed document based on its current state, eliminating the need to adapt the parameters of the LLMs or collect extensive training data for each app.

To validate its effectiveness, we tested our agent on  50 tasks across 10 different apps, ranging from social media and messaging to email, maps, shopping, and even complex image editing apps.   Both quantitative results and user studies underscore the advantages of our design, particularly its adaptability, user-friendliness, and efficient learning and operating capabilities across a wide range of applications. This underlines the potential of our agent as a versatile and effective tool in the realm of smartphone app operation.

In summary, this paper makes the following contributions:
 \begin{itemize}
\item  We open-source a multimodal agent framework,  focusing on operating smartphone applications with our developed action space.

\item We propose an innovative exploration strategy, which enables the agent to learn to use novel apps.

\item   Through extensive experiments across multiple apps, we validate the advantages of our framework, demonstrating its potential in the realm of AI-assisted smartphone app operation.

\end{itemize}

\begin{figure*}[t]
\centering
\includegraphics[width=1.0\textwidth]{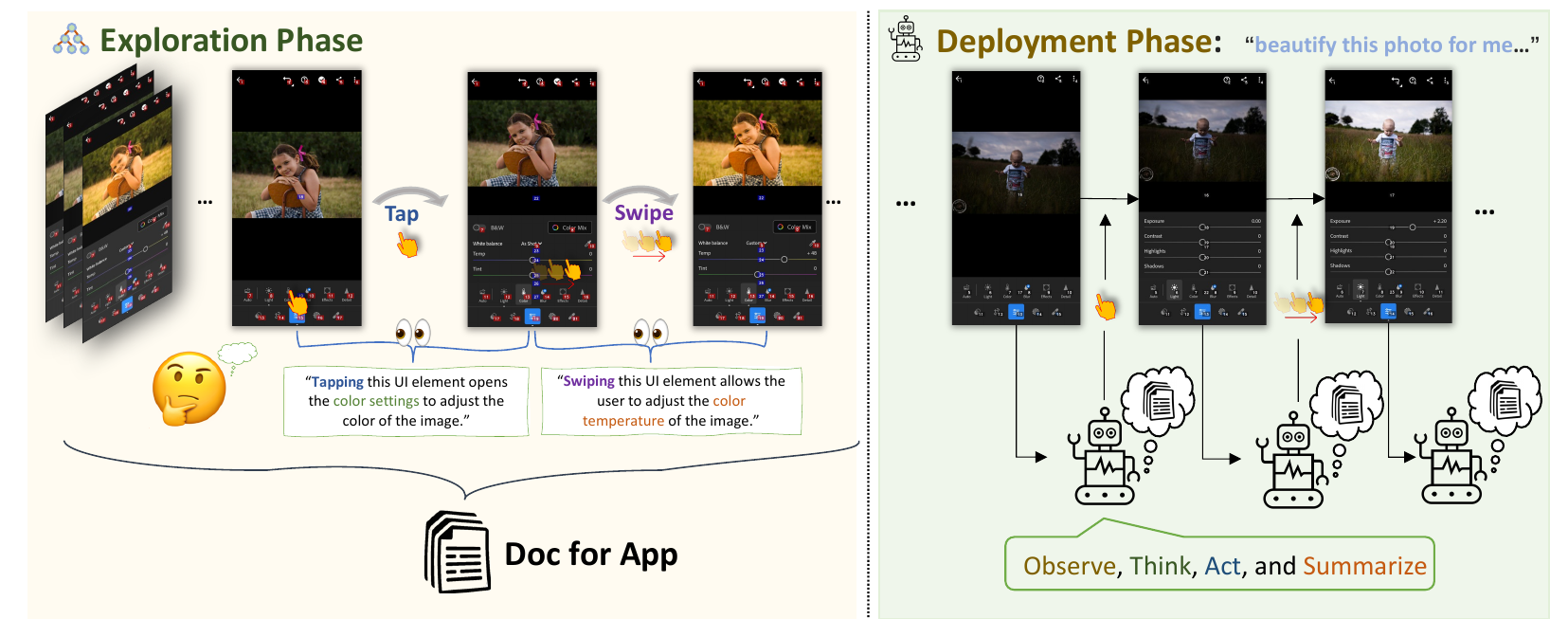}
\vspace{-0.5 em}
\caption{\textbf{Overview of our multimodal agent framework designed to operate smartphone applications.} The figure illustrates the two-phase approach of our framework. In the exploration phase, the agent interacts with a smartphone application and learns from their outcomes to create a comprehensive reference document. In the deployment phase, the agent utilizes the information compiled in this document to operate and navigate the apps effectively.
} 
\label{fig:main}
\vspace{-1.5em}
\end{figure*}

\section{Related Work}

\subsection{Large language models}
The development of ChatGPT~\cite{chatgpt} and GPT-4~\cite{gpt4} represents a crucial advancement in natural language processing. Unlike earlier large language models (LLMs), these new models~\cite{llama,llama2,glm,alpaca,vicuna} enable multi-round conversations and have the impressive ability to follow complex instructions. The integration of vision capabilities in GPT-4V~\cite{gpt4v} is a further milestone, enabling the language model to process and interpret visual data. This addition has broadened the scope of potential AI applications, allowing GPT-4 to undertake diverse tasks such as problem-solving, logical reasoning, tool usage, API calls, and coding.
Recent studies~\cite{yang2023dawn,yan2023gpt4v} have shown that GPT-4V can understand various types of images, including simple user interfaces (UIs) in popular smartphone apps.
However, challenges arise when the apps are new and their UIs are less typical, which highlights a major problem that our work aims to address.
Among open-source efforts from the industry and research community, the LLaMA series~\cite{llama,llama2} are the most popular equivalents and have been fine-tuned to acquire conversational abilities, employing a decoder-only architecture similar to ChatGPT~\cite{alpaca,vicuna}. Building upon LLaMA, many multimodal LLMs, such as LLaVA~\cite{llava,llava1.5}, ChartLlama~\cite{han2023chartllama}, and StableLLaVA~\cite{li2023stablellava}, also demonstrate vision understanding capabilities akin to those of GPT-4V. Nevertheless, a performance gap persists between these open-source models and GPT-4V, suggesting potential areas for further development.

\subsection{LLMs as agents}
The use of LLMs as agents for executing complex tasks has gained increasing attention.
Initiatives like AutoGPT~\cite{yang2023autogpt}, HuggingGPT~\cite{shen2023hugginggpt}, and MetaGPT~\cite{hong2023metagpt} illustrate this trend, and many projects demonstrate impressive capabilities, moving beyond basic language tasks to engaging in activities requiring higher cognitive functions, such as software development~\cite{qian2023chatdev,chen2021evaluating} and gaming~\cite{meta2022human,park2023generative,xu2023exploring}.
In this context, Yao~\emph{et al}.~\cite{yao2023react} introduce an innovative approach that synergizes reasoning and acting in LLMs, significantly enhancing their decision-making and interactive capabilities.
LLM-based agents are designed to utilize the advanced language and reasoning skills of LLMs to interact with and manipulate their environment~\cite{liu2023agentbench,gur2023realworld,xie2023openagents}. This includes performing tasks that require understanding context, making decisions, and learning from interactions~\cite{xi2023rise,hu2023language}.
Such agents are pivotal in applications where human-like cognitive abilities are essential.

The emergence of multimodal LLM agents~\cite{wang2023jarvis1,furuta2023multimodal,brohan2022rt,brohan2023rt-2,reed2022generalist}, capable of processing various inputs including text, images, audio, and video, has further broadened the scope of LLM applications.
This versatility is particularly beneficial for LLM-based agents, enabling them to interact more effectively with their environment and complete more complex tasks, be it completing household tasks in a physical world~\cite{saycan2022arxiv}, generating 3D assets via procedural tool use~\cite{sun20233d}, or mastering over 600 tasks across different domains at the same time~\cite{reed2022generalist}.
Our research contributes to this area by focusing on an agent designed to operate smartphone applications.
This agent's ability to interpret screenshots from the operating system demonstrates its flexibility and adaptability, making it a valuable tool in a wide range of applications.

\section{Method}
This section details the methodology behind our innovative multimodal agent framework. This framework enables an agent to interact with smartphone applications in a manner akin to human behavior. We first describe the experimental environment and action space, which are foundational elements of our system. Next, we discuss the exploration phase, where the agent learns app functionalities either through autonomous interactions or by observing human demonstrations. Finally, we outline the deployment phase, explaining how the agent applies its acquired knowledge to execute high-level tasks.

\subsection{ Environment and Action Space}
\textbf{Experimental Environment:} Our experimental environment is built on a command-line interface (CLI), allowing the agent to interact with smartphone apps. We chose the Android operating system for our experiments. The agent receives two key inputs: a real-time screenshot showing the app's interface and an XML file detailing the interactive elements. To enhance the agent's ability to identify and interact with these elements seamlessly, we assign each element a unique identifier. These identifiers are derived either from the resource ID in the XML file (if provided) or are constructed by combining the class name, size, and content of the element.
These elements are overlaid as semi-transparent numbers on the screenshot.
This helps the agent to interact accurately without needing to specify exact positions on the screen and enhances the agent's precision in controlling the phone.

\textbf{Action Space:} Our agent's action space mirrors common human interactions with smartphones: taps and swipes. We designed four basic functions:
 \begin{itemize}
\item  $\mathtt{ Tap(element: int):}$  This function simulates a tap on the UI element numbered on the screen. For example, $\mathtt{tap(5)}$ would tap the element labeled `5'.

\item $\mathtt{Long\_press(element: int):}$ This function emulates a long press (for 1 second) on a UI element.

\item  $\mathtt{ Swipe}$ $\mathtt{(}$ $\mathtt{element: int}$, $\mathtt{direction: str}$, $\mathtt{dist: str)}$: It allows the agent to swipe on an element in a specified direction (up, down, left, right) and distance (short, medium, long). For instance, $\mathtt{swipe(21, ``up", ``medium") } $ would swipe up on element `21' for a medium distance.

\item $\mathtt{ Text(text: str):}$ To bypass inefficient virtual keyboard typing, this function inputs text directly into an input field when a virtual keyboard is visible. For example, $\mathtt{text(``Hello, world!")}$ inputs the string ``Hello, world!".

\item $\mathtt{ Back( ):}$ A system-level function that helps the agent return to the previous UI page, especially useful for exiting irrelevant pages.

\item $\mathtt{ Exit( ):} $ A specialized function is employed to conclude processes, typically invoked upon successful task completion.

\end{itemize}

These predefined actions are designed to simplify the agent's interactions, particularly by eliminating the need for precise screen coordinates, which can pose challenges for language models in accurately predicting.

\subsection{Exploration Phase}

\textbf{Exploring by autonomous interactions.} The Exploration Phase is central to our framework. Here, the agent learns about the functionalities and features of smartphone apps through trial and error. In this phase, the agent is assigned a task and starts interacting autonomously with the UI elements. It uses different actions and observes the resulting changes in the app interface to understand how it works.
The agent, driven by a large language model, attempts to figure out the functions of UI elements and the effects of specific actions by analyzing screenshots before and after each action. This information is compiled into a document that records the effects of actions applied to different elements.
When a UI element is acted upon multiple times, the agent will update the document based on past documents and current observations to improve quality.
To make exploration more efficient, the agent stops further exploring UI elements if the current UI page seems unrelated to the main tasks of the app, like advertisement pages. In such cases, it uses the Android system's $\mathtt{ Back( )}$ function to return to the previous UI page. Compared with random exploration, such as Depth-First Search and Breadth-First Search, this goal-oriented exploration approach ensures that the agent focuses on elements crucial for the effective operation of the app. The agent also utilizes the LLM's existing knowledge about user interfaces to improve exploration efficiency. The exploration stops when the agent completes the assigned task.

\textbf{Exploring by watching demos.} An alternative and often more effective exploration method involves the agent observing human demonstrations. These demonstrations provide the agent with examples of efficient app usage, especially for understanding complex functionalities that might be challenging to discover through autonomous interactions. In this method, a human user operates the apps while the agent observes, recording only the elements and actions employed by the human. This strategy narrows down the exploration space and prevents the agent from engaging with irrelevant app pages, making it a more streamlined and efficient approach compared to autonomous interactions.

\subsection{Deployment Phase}
Following the exploration phase, the agent is well-equipped to execute complex tasks based on its accrued experience. The agent adheres to a step-by-step approach when given a task, with each step encompassing access to a screenshot of the current UI and a dynamically generated document detailing the functions of UI elements and the actions' effects on the current UI page.
The prompts also provide detailed explanations of all available actions.
In each step, the agent is first tasked with providing its observations of the current UI, followed by articulating its thought process concerning the task and current observations. Subsequently, the agent proceeds to execute actions by invoking available functions. 
After each action, the agent summarizes the interaction history and the actions taken during the current step. This information is 
incorporated into the next prompt, which provides the agent with a form of memory.
This meticulous approach enhances the reliability and interpretability of the agent's actions, thereby facilitating more informed decision-making. The deployment phase stops when the agent determines that the task has been accomplished, at which point it can exit the process by taking the $\mathtt{ Exit( )} $ action.

\begin{figure*}[t]
\centering
\includegraphics[width=0.99\textwidth]{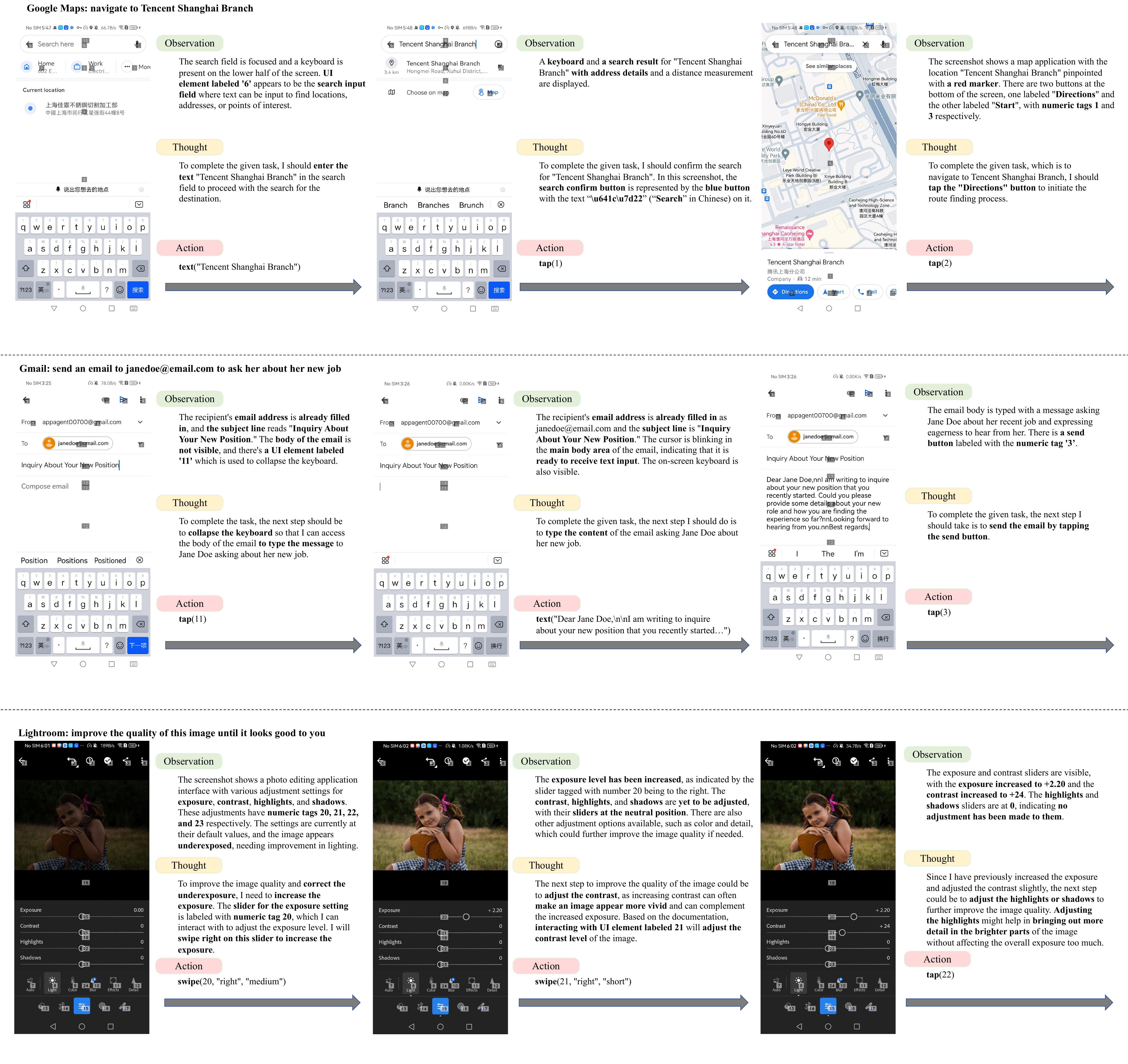}
\vspace{-0.5 em}
\caption{\textbf{Qualitative Task Evaluation Across Three Apps. } This figure presents qualitative results for three distinct tasks conducted on Google Maps, Gmail, and Lightroom. It showcases AppAgent's ability to accurately perceive, reason, and execute tasks, demonstrating its competence in various application contexts. Due to space constraints, some less critical details have been omitted from the description.
} 
\label{fig:qualatative}
\vspace{-1.5em}
\end{figure*}

\begin{table*}[t]
\centering
\begin{tabular}{llllll}
\toprule[1pt]
Method                                    & Document            & Action Space & SR $\uparrow$ & Reward $\uparrow$& Avg. Steps \\ \toprule[1pt]
\multicolumn{1}{l}{\multirow{2}{*}{GPT4 (Baseline)}} & None           & Raw          & 2.2\%  & 0.6     & 4.0          \\ \cline{2-6} 
\multicolumn{1}{c}{}                      & None           & \textbf{Ours}         & 48.9\%  & 3.5      & 6.9          \\ \toprule[0.6pt]
\multirow{3}{*}{\textbf{AppAgent}}                     & \textbf{Auto. Exploration}     & \textbf{Ours}          & 73.3\%  & 5.1      & 4.4         \\ \cline{2-6} 
                                          &\textbf{Watching Demos }          & \textbf{Ours}          & 84.4\% & 4.7     & 5.1         \\ \cline{2-6} 
                                          & \textbf{Manually Crafted } & \textbf{Ours}          & 95.6\% & 5.5     & 5.5         \\ \toprule[1pt]
\end{tabular}
\caption{\textbf{Evaluating Design Choices in AppAgent Performance.} This table contrasts different design elements within AppAgent. Key findings include: our custom-developed action space surpasses the raw action space in efficiency; the exploration phase, incorporating both autonomous interaction and observation of human demonstrations, significantly enhances agent performance; and the auto-generated documentation yields outcomes on par with those derived from manually crafted documents. }
\label{tab_design}
\end{table*}

\begin{table*}[t]
\centering
\begin{tabular}{lllll}
\toprule[1pt]
Method                                    & Document            & Action Space &  Avg. Rank $\downarrow$ & Num. Tools   \\ \toprule[1pt]
GPT4 (Baseline)        & None           & \textbf{Ours}         & 2.30      & 2.4          \\ \toprule[0.6pt]
\multirow{2}{*}{\textbf{AppAgent}}                 &\textbf{Watching Demos }          & \textbf{Ours}          & 1.95      &5.8          \\ \cline{2-5} 
                                          & \textbf{Manually Crafted } & \textbf{Ours}          & 1.75      & 4.0           \\ \toprule[1pt]
\end{tabular}
\caption{\textbf{Case study on image editing tasks with Lightroom App.} We conduct a user study to rank the image editing results of different methods. Our agents produce better results than the GPT-4 baseline.  }
\label{tab_case_study}
\end{table*}

\section{Experiments}
In this section, we will present our evaluation of the multimodal agent framework through a combination of quantitative and qualitative experiments. Our primary goal is to assess the agent's performance and its ability to operate a diverse set of smartphone applications effectively.

\subsection{Experimental Setup}
To comprehensively evaluate our method, we construct a benchmark that includes 10 popular applications, each serving various purposes. These applications include Google Maps, Twitter, Telegram, YouTube, Spotify, Yelp, Gmail, TEMU, Clock, and Lightroom. We have intentionally chosen this diverse set of apps to test the agent's adaptability across various functions and interfaces.
In particular, to gain a more comprehensive insight into the vision capabilities of our agent, we conducted an in-depth case study using Adobe Lightroom, an image-editing application. This specific case study allowed us to evaluate the agent's proficiency in handling visual tasks and its ability to interpret and manipulate images within the app.
For the exploration phase, we capped the maximum number of steps at 40. During testing, we limited the maximum number of steps to 10.
For these experiments, we utilized the state-of-the-art multimodal large language model, GPT-4. GPT-4 is equipped to process interleaved image-and-text inputs effectively. This unique capability enables our agent to interpret and interact with both visual and textual information seamlessly within the applications.

\subsection{Design and Analysis}
\textbf{Baselines.} To comprehensively evaluate our multimodal agent framework, we considered various design choices and their impact on performance. We conducted experiments using different configurations to provide valuable insights into the agent's behavior.
We started with \textbf{GPT-4} without any reference documents during testing and examined its performance both with the raw action API and our simplified action space.
Next, we explored different ways to generate guiding documents for the agent. These included documents generated through autonomous exploration, watching human demonstrations, and the manually crafted document as an oracle benchmark.

 To effectively compare the performance of different methods, we employed three key metrics:\\
\textbf{Successful Rate (SR):} This metric measures the average rate at which the agent successfully completes tasks within an app.  If the agent fails to finish the task in 10 steps, it is considered a failure.\\
\textbf{Reward:} To provide a more fine-grained measurement, we developed a reward model to assess performance. For each task within an app, we scored different UI pages. The closer the UI page was to the objective, the higher the score received. This means that even if the agent failed to complete the task, it would still receive credit based on its final state.\\
\textbf{Average Steps:} We also reported the average number of steps required to successfully finish tasks across the selected applications.

\textbf{Results.} The comparison of our experimental results is presented in Table~\ref{tab_design}. 
We report the average performance of  45 tasks on 9 of the 10 previously described apps. Notably, we excluded Lightroom from this evaluation, as assessing task completion in this application presented inherent ambiguities.
As demonstrated, our simplified action space significantly improves the performance of the GPT-4 baseline. Our observations indicate that LLM struggles with producing accurate xy coordinates, while our simplified action space eliminates this challenging requirement.
Additionally, documents generated through autonomous exploration and observing human demonstrations proved to be highly effective. Their results consistently outperformed the GPT-4 baseline and are comparable to the results of human-written documents, which highlights the efficacy of our design in enhancing the agent's performance across a diverse set of applications.

\textbf{Qualitative results.}
In Fig.~\ref{fig:qualatative}, we provide examples showcasing the agent's execution process for various tasks. This qualitative analysis serves to demonstrate the agent's capacity to accurately perceive, reason, and act in response to given tasks. For a more comprehensive understanding of our agent's capabilities, please refer to our project page, which includes additional demonstration videos.

\subsection{Case Study}

To gain deeper insights into the vision capabilities of our agent, we conducted an extensive case study using Adobe Lightroom, an image-editing application. This specific case study allowed us to evaluate the agent's proficiency in handling visual tasks, which was previously impossible for text-only agent models.
Lightroom, as an image-editing app with various editing tools, demands a wide range of operations, such as selecting appropriate tools and manipulating image parameters. This case study provides a robust evaluation of the agent's overall capabilities. Additionally, the open-ended nature of image editing tasks allows us to assess the agent's problem-solving abilities.
We prepared five images with visual issues, such as low contrast and overexposure. Various variants of our model, as previously illustrated, were used to edit these images. A user study was conducted to rank the editing results produced by different methods.
We also reported the average number of tools used for image editing, providing an additional reference to the editing process's complexity.
All models were assigned the task of ``fix this image until it looks good to you'' without specifying the image's problems. The comparison of the results is presented in Table~\ref{tab_case_study}. As we can see, our agent model with documents yields consistently better results than the GPT-4 baseline, which emphasizes the influence of documents in our design.
The generated documents by watching the demonstration produced comparable results with the results of manually crafted documents, which suggests the effectiveness of the exploration phase. We also find that with a document, the agent tends to use various tools to improve the image quality, while the GPT-4 baseline uses fewer tools.

\section{Conclusion}
In this paper, we have introduced a novel multimodal agent framework that leverages the vision capabilities of large language models to operate smartphone applications in a human-like manner.  Our approach eliminates the need for system back-end access and offers security, adaptability, and flexibility advantages.  Our exploration-based learning strategy allows the agent to quickly adapt to new applications with unfamiliar user interfaces, making it a versatile tool for various tasks. Our extensive experiments across various apps highlight our agent's ability to handle diverse high-level tasks and underscore its adaptability and learning efficiency.

\textbf{Limitation.} We have adopted a simplified action space for smartphone operations, which means that advanced controls such as multi-touch and irregular gestures are not supported. This limitation may restrict the agent's applicability in some challenging scenarios. Nevertheless, we recognize this as an avenue for future research and development.

\bibliography{anthology,custom}
\bibliographystyle{acl_natbib}

\end{document}